\documentclass[letterpaper, 10 pt, conference]{ieeeconf}  

\IEEEoverridecommandlockouts                              

\usepackage{algorithm}
\usepackage{algorithmic}
\usepackage{amsmath} 
\usepackage{amssymb} 
\usepackage{booktabs} 
\usepackage{multirow} 
\usepackage{graphicx}
\let\labelindent\relax
\usepackage[inline]{enumitem}
\usepackage{cite} 
\usepackage[breaklinks=true]{hyperref} 
\hypersetup{
    colorlinks=true,
    linkcolor=black,      
    citecolor=black,      
    urlcolor=black,
    bookmarksnumbered=true,
    bookmarksopen=true
}
\usepackage{url}
\usepackage{xcolor}
\usepackage{caption}

\title{\LARGE \bf
TWISTED-RL: Hierarchical Skilled Agents for Knot-Tying without Human Demonstrations
}

\author{Guy Freund$^{1}$, Tom Jurgenson$^{2}$, Matan Sudry$^{2}$, and Erez Karpas$^{2,3}$\\
{\small $^{1}$Reichman University, $^{2}$Technion – Israel Institute of Technology, $^{3}$The University of Texas at Austin}%
\thanks{Correspondence to: \href{mailto:guyfreund@gmail.com}{\tt guyfreund@gmail.com}}
\thanks{Project Page: \href{https://sites.google.com/view/twisted-rl}{\tt sites.google.com/view/twisted-rl}}
}

\begin{document}

\maketitle
\thispagestyle{empty}
\pagestyle{empty}

\begin{abstract}

Robotic knot-tying represents a fundamental challenge in robotics due to the complex interactions between deformable objects and strict topological constraints. We present TWISTED-RL, a framework that improves upon the previous state-of-the-art in \textit{demonstration-free} knot-tying (TWISTED), which smartly decomposed a single knot-tying problem into manageable subproblems, each addressed by a specialized agent. Our approach replaces TWISTED's single-step inverse model that was learned via supervised learning with a multi-step Reinforcement Learning policy conditioned on abstract topological actions rather than goal states. This change allows more delicate topological state transitions while avoiding costly and ineffective data collection protocols, thus enabling better generalization across diverse knot configurations. Experimental results demonstrate that TWISTED-RL manages to solve previously unattainable knots of higher complexity, including commonly used knots such as the Figure-8 and the Overhand. Furthermore, the increase in success rates and drop in planning time establishes TWISTED-RL as the new state-of-the-art in robotic knot-tying without human demonstrations.

\end{abstract}

\begin{figure*}[!t]
    \centering
    \includegraphics[width=\textwidth]{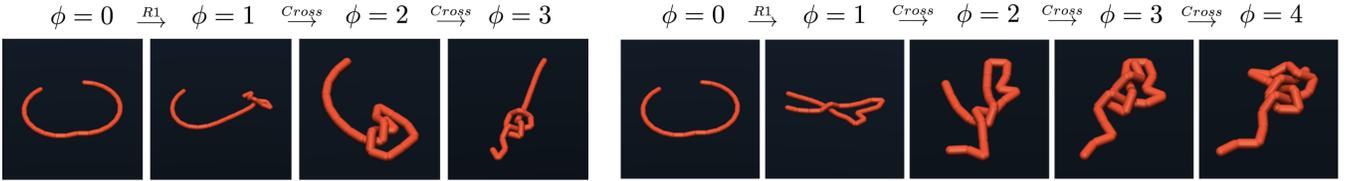}
    \caption{The bottom row shows the Figure-8 knot (left) and the Overhand knot (right) performed by TWISTED-RL. The top row shows the complexity of the rope measured in crossing number, and the transitions are labeled with the corresponding Reidemeister moves.}
    \vspace{-1em}
    \label{fig:trajectories}
\end{figure*}

\section{Introduction}
\label{sec:intro}

Robotic knot-tying represents a uniquely challenging task within robotics due to the inherent complexity arising from deformable objects, high-dimensional state spaces, and stringent topological constraints~\cite{zhu2022challenges,gu2023survey}. Unlike rigid-object manipulation, ropes and threads do not maintain a fixed structure, making their state representation and control significantly more challenging. Real-world applications underscore the importance of solving this problem effectively; domains such as robotic surgery~\cite{wang2010suturing}, textile manufacturing~\cite{6631059}, and cable management in industrial setups~\cite{shivakumar2023sgtm}, rely heavily on robust and reliable knot-tying capabilities. The complexity of robotic knot-tying manifests in several key difficulties. First, the topological state space -- which captures the essential structure of knots while abstracting away geometric details -- grows exponentially with the complexity of the knot, making exhaustive search approaches computationally infeasible. Second, collecting consistent data for supervised learning approaches is particularly challenging because manipulating tangled ropes without a performant policy tends to untangle the rope and remove all complexity, making it unlikely to observe complex manipulation events in the data~\cite{sudry2023hierarchical}. To overcome these limitations, we use reinforcement learning (RL), an approach that iteratively improves the policy while being exposed to increasingly complex data as the learning process continues. However, this introduces a third challenge: the reward landscape in knot-tying is extremely sparse, as successful topological transitions are rare occurrences amid countless failed attempts, creating a significant impediment to effective learning. To better understand these challenges, we must consider the fundamental nature of the knot-tying task, as demonstrated in Figure~\ref{fig:trajectories}. The robotic knot-tying problem involves transforming an initially unstructured rope into a target knot. This process can be viewed at two distinct yet interconnected levels: a high-level representation that describes the rope in terms of its topological state -- essentially which segments cross over or under other segments, forming the knot's mathematical identity regardless of how it is stretched or positioned (e.g., the type of knot or its intermediate form), and a low-level representation that captures the precise geometric configuration of the rope in physical space (e.g., 3D coordinates and orientations). At the high level, knot-tying can be described through a series of abstract, topological moves -- commonly known as Reidemeister moves~\cite{reidemeister1983knot} -- that progressively guide the rope toward the desired knot structure. Each high-level move encapsulates changes to the overall topology of the rope without specifying the exact geometric details required for execution. The low-level challenge lies in translating these abstract topological moves into precise physical manipulations of the rope. Specifically, the robot must execute a sequence of detailed, geometry-aware actions, such as moving or bending the rope, to achieve each high-level transition. The current state-of-the-art in demonstration-free knot-tying, TWISTED~\cite{sudry2023hierarchical}, is a hierarchical planning framework for robotic knot-tying that unifies abstract topological reasoning with low-level geometric execution using an inverse model dependent on the goal topological state, which is trained without human-demonstrations. While it demonstrates effective performance in complex knot-tying tasks, its low-level policy is subject to three key limitations:
\begin{itemize}
    \item Single-Step Execution:
    The low-level policy, was trained and used to traverse edges between high-level states in one shot, that is, a solution of TWISTED is a series of robot actions that each \emph{increase} the complexity of the knot. This nature of the solution is very restrictive, preventing the system to "tweak" intermediate states within the same topological state, that might be easier to be manipulated into higher-level states.

    \item State-Goal Conditioning: The action prediction model is conditioned on the \textit{next topological state}, whose representation size grows with the complexity of the knot. Such a sparse and discrete representation would prevent the agent from learning, as many states would not be present in the fixed training data collected by TWISTED, making predictions for such states arbitrary.

    \item Data Collection: TWISTED relies on a manually-defined (yet automated) data collection flow. This becomes a severe bottleneck when trying to collect data of complex knots. For example, TWISTED notes that obtaining a complex dataset requires 1 hour to collect 537 transitions on a single CPU core.
\end{itemize}

In this work, we present \textit{TWISTED-RL}, which addresses these limitations, establishing a major advancement in knot-tying without demonstrations. Our contributions can be summarized as follows:

\begin{itemize}
    \item Multi-step Low-level Policies for Topological Transitions: We use RL policies capable of executing multi-step sequences of low-level actions, significantly improving the capability to perform complex topological transformations, by making delicate adjustments to the rope. 
    Moreover, by using the RL framework our agent improves as more data is collected, eliminating the need for complicated data collection procedures.
    
    \item Transferable Action-Conditioned Skills: Unlike TWISTED state-conditioned goal representation, our policies goals are  the abstract Reidemeister moves (topological actions), yielding transferable skills that generalize across different levels of complexity.

    \item Improved Performance on Complex Knots:
    Experimental evaluation demonstrates superior performance of TWSITED-RL compared to TWISTED on knots with high crossing numbers, achieving substantial improvements in task success rates (often enabling previously unattainable knots) and runtime efficiency compared to TWISTED. To the best of our knowledge, TWISTED remains the only existing framework that tackles complex robotic knot-tying in a completely demonstration-free manner, making it the most relevant and direct baseline for evaluating our contributions.
\end{itemize}

\section{Background}
\label{sec:background}

The robotic knot-tying problem can be formalized as a hierarchical task operating on two levels\footnote{For a full formal task description, see Sec.~\ref{appdx:extended-env} in the appendix.} \cite{sudry2023hierarchical}. At the high level, a discrete space $\mathcal{S}$ represents topological configurations of the rope, independent of geometric details. $\mathcal{S}$ is characterized by the crossing number $\phi:\mathcal{S}\rightarrow\mathbb{N}$, a complexity measure defined as the minimum number of crossings in any representation of a topological state. At the low level, a continuous space $Q$ represents physical rope configurations. To connect both levels of representation, a function $Top: Q \rightarrow \mathcal{S}$ maps physical configurations to their corresponding topological states. The low-level action space $C$ consists of continuous curve-based motion primitives, where each action $c \in C$ is parameterized by the rope link to grasp, a planar target position, and a maximum lift height, defining a trajectory for the robot manipulator. The knot-tying challenge lies in transforming an initial rope configuration $q_0 \in Q$ with $Top(q_0) = S_0$ to a target topological state $S^g$. This requires planning at the topological level -- finding a sequence of valid Reidemeister moves $A_0, A_1, \ldots, A_{m-1} \in \mathcal{A}$ that transform $S_0$ to $S^g$ -- and then produce a sequence of low-level robot actions $c \in C$ that successfully execute the transitions in the high-level plan, effectively translating abstract topological moves into physically realizable manipulations of the rope.

\subsection{TWISTED}

TWISTED's~\cite{sudry2023hierarchical} high-level planner begins from the initial topological state $S_0$ (called the unknot where the no rope segment crosses another) and searches in the high-level graph $G=(\mathcal{S},\mathcal{A})$ for a path to a desired goal state $S^g$ representing a target knot. This search yields a sequence of topological actions $A_0, A_1, \dots, A_m$ and corresponding abstract states $S_0, S_1, \dots, S_m,S_{m+1}=S^g$. Each action $A_i$ is a Reidemeister move meant to increase the crossing number and progress toward the goal. The resulting high-level plan provides a coarse template, but its execution must be grounded in real-world physics. To execute these abstract moves, TWISTED employs a learned low-level inverse model that maps each desired topological transition $(S \rightarrow S')$ into \textit{a single} continuous action $c \in C$. The low-level action is then applied in simulation using a physics-based rope model $f$, resulting in a new geometric state $q' = f(q, c)$. If the resulting state corresponds to the expected topological successor $S' = Top(q')$, the plan proceeds; else if the crossing number of the resulting state is bigger than $\phi(S')$, it adds it to the graph of reachable configurations so far, named $T$. Otherwise, TWISTED falls back to re-planning, retrying alternate high-level paths from a different low-level state in $T$. The inverse model bridges topological planning and physical execution, translating desired high-level transitions into concrete robot actions that can be physically realized. It is goal-conditioned: given the initial configuration $q$ and a desired topological state $S'$, it predicts the most likely curve $c$ to achieve that transition. It is trained via supervised learning on a dataset of sampled transitions, generated by applying random actions and retaining those that yield valid topological changes. Together, TWISTED combines a symbolic high-level planner with a continuous stochastic low-level policy to construct knots incrementally. The high-level planner encodes the space of topological possibilities, while the inverse model grounds these abstract moves in executable robot actions. To improve robustness, it performs multiple iterations of planning and execution, allowing the system to recover from errors until a solution is found. Moreover, a small probability of random exploration ensures that the agent does not get stuck in local optima during execution and that a solution will always be found with unlimited runtime.

\subsection{Reinforcement Learning (RL)}
\label{sec:rl}

RL\cite{712192} is a machine learning paradigm focused on training agents to make sequential decisions in an environment to maximize cumulative rewards. For robotic manipulation tasks like knot-tying, goal-conditioned RL is particularly relevant. In this framework, an agent interacts with the environment by taking actions based on its current state and a specified goal, receiving feedback that measures progress toward that goal. This is formalized as a goal-conditioned Markov decision process (GC-MDP), characterized by a tuple $(S, A, f, \mathcal{R}, \mathcal{G}, \gamma)$ where $S$ is the state space, $A$ is the action space, $f: S \times A \longrightarrow S$ is the transition function, $\mathcal{G}$ is the goal space, $\mathcal{R}: S \times S \times \mathcal{G} \longrightarrow \mathbb{R}$ is the reward function that evaluates transitions with respect to goals, and $\gamma$ is the discount factor. The agent's objective is to learn a policy $\pi: S \times \mathcal{G} \longrightarrow A$ that maximizes the expected cumulative discounted reward for any given goal, i.e., \(\pi^* = \arg\max_{\pi} \mathbb{E} \left[ \sum_{t=0}^{\infty} \gamma^t r_t \,\middle|\, \pi, g \right]\). Our method uses the commonly-used Soft Actor Critic (SAC)~\cite{sac} RL algorithm that has demonstrated robust convergence properties and good sample efficiency in a variety of continuous control benchmarks.

\begin{figure}[t]
    \centering
    \includegraphics[width=\columnwidth]{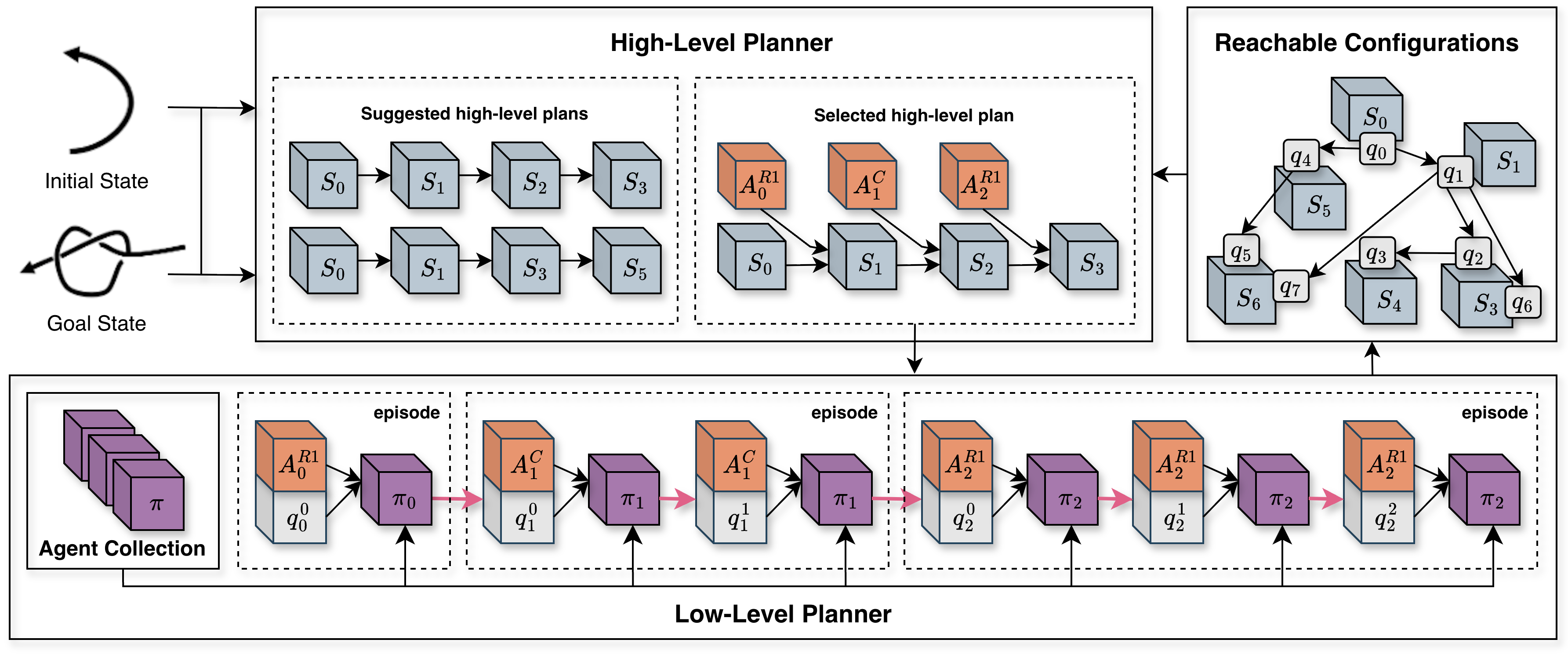} 
    \caption{Illustration of TWISTED-RL System Architecture. Top row: the unmodified TWISTED components showing the reachable configurations and high-level planning. Bottom row: TWISTED-RL low-level execution with three episodes showing progression between crossing numbers. First episode: $0\rightarrow1$ with R1 action, second episode: $1\rightarrow2$ with Cross action, third episode: $2\rightarrow3$ with R1 action, each using its specialized agent. Red arrows indicate MuJoCo simulation steps. \vspace{-1.5em}}
    \label{fig:twisted-rl}
\end{figure}

\section{Problem Formulation}
\label{sec:problem-formulation}

In our work we focus on the particular part of the knot-tying problem of transitioning effectively between subsequent high-level states.
We formally define this problem of \textit{low-level} robotic knot-tying as a GC-MDP, which is described by the tuple $(Q, C, f, \mathcal{R}, \mathcal{G}, \gamma)$ where $Q$ is the state space representing the continuous, high-dimensional configuration space of the rope; $C$ is the action space defining the curve-based motion primitives that approximate abstract Reidemeister moves; $f: Q \times C \rightarrow Q$ is the transition function modeling the physics-based rope dynamics implemented through the simulator; $\mathcal{G}$ is the goal space representing desired high-level transitions between topological states, with each goal $g \in \mathcal{G}$ corresponding to a specific transition $(S \overset{A}{\longrightarrow} S^g)$ in the high-level graph
; the reward function $\mathcal{R}: Q \times Q \times \mathcal{G} \rightarrow \mathbb{R}$ assigns +1 for achieving the goal state $S^g$, 0 for remaining in the current state $S$, and -1 for any other state transition\footnote{While attributing the same penalty to all non-successful transitions has the same optimal policy, it has an unfavorable local policy that prefers to terminate episodes early in order to avoid accumulating negative rewards. This discourages the agent from exploring. We opted for a simple setup, and leave investigation of alternative reward formulations to future work.}; and the discount factor $\gamma$ determines the relative importance of immediate versus future rewards. The objective is to learn a low-level policy \(\pi: Q \times \mathcal{G} \longrightarrow C\) that reliably executes goal topological transitions. Specifically, given an initial configuration \(q \in Q\) corresponding to a high-level state \(S = Top(q)\) and a desired high-level transition $( S \overset{A}{\longrightarrow} S^g)$, the policy \(\pi\) should determine a sequence of curves that transform the rope into a configuration corresponding to the target knot \(S^g\).

\section{Method}
\label{sec:method}

TWISTED-RL builds upon the hierarchical structure of TWISTED but uses RL at the low level to overcome TWISTED's core limitations (see Figure~\ref{fig:twisted-rl}). While TWISTED executes each high-level topological move using a single predicted low-level curve, our method instead employs multi-step episodes that may comprise multiple low-level actions. This approach enables the policy to perform fine adjustments, isolate rope segments, and robustly achieve transitions that would otherwise fail under a single-step regime. The high-level planning layer of TWISTED-RL remains identical to TWISTED, while the execution of each high-level move is delegated to a learned multi-step policy trained with an off-the-shelf algorithm for RL (SAC), which begins \textit{tabula-rasa} with no additional data from TWISTED\footnote{For additional details regarding our full training scheme, see Sec.~\ref{sec:training-appendix} in the appendix.}. Additionally, our policy is conditioned on the high-level action rather than the goal topological state, enabling the formation of reusable "skills" -- e.g., an "R1 skill" that allows for generalizable, skill-based behaviors such as "how to perform an R1 move". In the following sections, we detail each component of our learning framework: we begin by describing our training scheme of multiple SAC agents -- based upon the high-level action (R1, R2, and Cross), the crossing number $n\in \mathbb{N}$, both or neither. We also describe how they are integrated into TWISTED's high-level planner (Sec.~\ref{sec:high-level-planner}). We explain how each policy is conditioned on high-level actions as skills (Sec.~\ref{sec:skill_conditioned}), outline the multi-step structure of low-level episodes (Sec.~\ref{sec:multi-step-episodes}). Finally, we detail our goal and initial state selection strategies for robust training (Sec.~\ref{sec:selection}).

\subsection{From Monolithic Policies to Specialized Agents}
\label{sec:high-level-planner}
TWISTED-RL groups all transitions $(S \overset{A}{\longrightarrow} S')$ into sets according to the high-level action type of $A$, the crossing number of $S$ ($\phi(S)$) or both of these. We also explore grouping all transitions into a single set, yielding 4 different variants of our method: (1) \emph{TWISTED-RL-G}, a single agent trained on the full high-level graph $G$; (2) \emph{TWISTED-RL-A}, which employs specialized agents for each high-level action type (R1, R2, Cross); (3) \emph{TWISTED-RL-C}, which uses a single agent for each crossing number; and (4) \emph{TWISTED-RL-AC}, which employs specialized agents for each combination of high-level action type and crossing number. This approach focuses on learning independent policies for specific sub-tasks rather than a monolithic policy, as in TWISTED, which used a single inverse model to handle all high-level transitions, leading to significant challenges in data-collection and data-balancing.
Our specialized training approach is motivated by critical data-balancing considerations. Different combinations of crossing numbers and high-level actions vary significantly in complexity and difficulty, naturally resulting in substantial imbalances in data availability and quality. By training dedicated policies for each set, we address these imbalances explicitly, ensuring robust and balanced data coverage for each specific sub-problem. On the other hand, fewer sets enhance the generalization capabilities of the policies, and we explore this tradeoff in our experiments. During inference, given a target knot, TWISTED's high-level planner determines a high-level plan to reach the goal. While following this plan, based on the intended high-level transition, the system selects the appropriate specialized agent to perform the needed high-level transition, as illustrated in the bottom row of Fig.~\ref{fig:twisted-rl}. 

\subsection{Learning Policies Dependent on High-Level Action Skills}
\label{sec:skill_conditioned}

We hypothesize (and empirically demonstrate) that developing \emph{action-conditioned} skills -- policies that learn to execute parameterized Reidemeister moves -- yields greater robustness and transferability than conditioning on goal topological states. Specifically, we condition the policy on the high-level action \(A \in \mathcal{A}\), encoding its parameters alongside the low-level state \(q\) (Figure~\ref{fig:twisted-rl}).
This design differs from prior works such as TWISTED, which trained policies conditioned on \emph{goal states}. While goal-based conditioning can work in simpler settings, it suffers from two key limitations:
(1) As the crossing number increases, the topological state representation (P-Data~\cite{pdata}) grows in size, requiring variable-length representations that cannot be accommodated in a fixed neural network architecture, even when the same fundamental action (e.g., R1) is required.
(2) State goal conditioning prevents skill reuse across knots, as each goal is treated as a separate entity rather than an abstraction. In contrast, our approach promotes generalization by learning skills tied to topological \emph{transformations} rather than \emph{configurations}. E.g., an “R1 skill” learned on simple knots can transfer to more complex settings, since the fundamental mechanics of the R1 move remain consistent. Concretely, given a goal high-level transition $g = (S \overset{A}{\longrightarrow} S^g)$, we encode the action \(A\) into a fixed-length vector \(\rho(g)\) that captures the parameters of the corresponding Reidemeister move.
For the low-level state representation, we apply a projection function $\eta: Q \rightarrow P$ that extracts only the positional information $p$ from the full state $q=(p,o)$, effectively omitting the orientation components $o$. This simplification reduces input dimensionality while retaining the essential geometric information needed for manipulation. The action encoding $\rho(g)$ is then concatenated with the position representation $\eta(q)$, forming the complete input to our policy:
\(
\pi: \eta(Q) \times \rho(\mathcal{G}) \longrightarrow C
\).

\subsection{Multi-Step Episodes}
\label{sec:multi-step-episodes}

TWISTED models each goal high-level topological transition as a single-step episode, where a low-level policy executes a single curve to induce the desired change. From a control perspective, this could be considered a \emph{contextual bandit-style} approach, which treats each move as atomic and attempts to learn a one-shot mapping from state to action. However, we argue that this formulation is overly restrictive for complex knots, where the crossing number increases. In such settings, transitions often require intermediate "adjustment" actions: steps that isolate a specific rope segment or reposition parts of the rope to enable a successful topological move. Inspired by how humans naturally perform knot-tying through sequences of deliberate sub-actions, we instead frame each topological transition as a multi-step \emph{episode}, allowing the policy to execute a sequence of up to $M$ curves. The episode terminates when either the topological state changes (whether to the goal state or an unintended state) or after reaching the maximum allowed steps $M$. Importantly, the agent can remain in the initial high-level state during the episode to perform necessary geometric adjustments before executing the transition. This multi-step formulation provides several advantages. It enables more complex manipulations by breaking them into manageable sub-steps, allows for recovery from minor errors, and more closely mimics humans, which rarely achieve complex transitions in a single motion.

\subsection{Goal and Initial Low-Level State Selection}
\label{sec:selection}

Inspired by prior work on exploration-driven RL~\cite{goexplore}, our goal selection strategy during training maximizes coverage over all high-level transitions. To achieve this, we \emph{randomly} select a high-level goal transition $(S \overset{A}{\longrightarrow} S^g)$.  Given the selected initial high-level state \( S \), the initial low-level state \( q \in Q \) is sampled uniformly from all low-level states that satisfy \( Top(q) = S \) that were encountered during training thus far. This uniform sampling ensures robustness across a variety of geometric configurations, which supports generalization by training the agent under diverse initial conditions. This randomized selection approach serves our objective of exploration and coverage in two key ways. First, by sampling uniformly across all possible goal states, we ensure that the agent eventually attempts all valid transitions. Second, the randomization naturally prevents the agent from getting stuck repeatedly attempting the same challenging transitions that it currently performs poorly on, which could otherwise create a negative learning cycle. For more details, see Appendix Sec.~\ref{sec:additional-details-appendix}.

\section{Experiments}
\label{sec:experiments}

\begin{figure*}[!t]
    \centering
    \includegraphics[width=\textwidth]{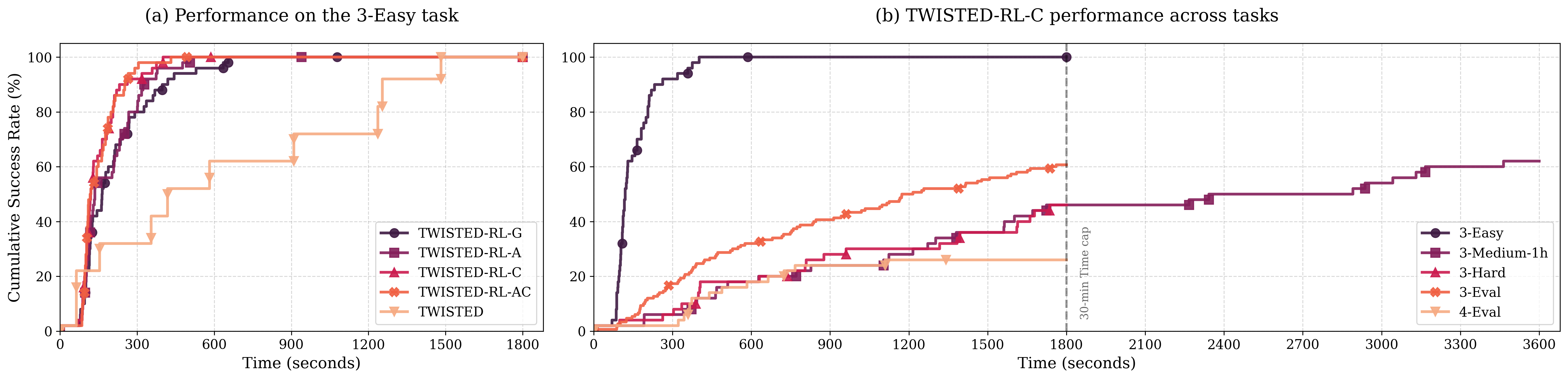}
    \caption{Left: Anytime performance analysis of TWISTED and TWISTED-RL variants on the 3-Easy test set. Right: Anytime performance analysis of TWISTED-RL-C across all test sets. The X-axis is the runtime in seconds, and the Y-axis is the cumulative success rate (\%).
    \vspace{-1.5em}}
    \label{fig:runtime}
\end{figure*}

In this section, we evaluate TWISTED-RL variants against the state-of-the-art in demonstration free knot-tying -- TWISTED. We explore the performance of these methods on different sets of knots and evaluate the tradeoff between different data-partitions of the high-level transitions.

\subsection{Experimental Setup}
We evaluate TWISTED-RL against TWISTED as our primary baseline, as it represents the only existing method for demonstration-free robotic knot-tying that addresses the same problem formulation. TWISTED established the first successful approach to learning complex knot-tying without demonstrations and demonstrated superior performance compared to alternative methods in its original evaluation. Therefore, establishing improvements over TWISTED provides strong evidence of our method's effectiveness and represents a meaningful advancement in the field. Our experiments use a MuJoCo-based physics simulation environment~\cite{mujoco} consistent with TWISTED~\cite{sudry2023hierarchical}. All experiments were conducted on standard computational hardware and repeated across 5 different random seeds to account for stochasticity. We imposed a 30-minute time cap per goal state for all test sets, except the 3-Medium-1h set, where the allowed runtime is 1 hour. We evaluate on several carefully designed test sets of knots. First, the standard 3-crossing sets consist of states with 3 crossing numbers of medium (3-Medium) to hard (3-Hard) complexity that were randomly selected following the protocol established in previous work~\cite{sudry2023hierarchical}. The 3-Medium complexity states appear 120-130 times in TWISTED's dataset, corresponding to the 33rd percentile of frequency. The 3-Hard complexity states appear only once, corresponding to the 1st percentile (representing the most challenging cases). Second, we introduce an Extended 3-Crossing Set (3-Eval), comprising a more diverse collection of 30 states with crossing number 3. These states span a broader range of difficulty levels, based on their frequency distribution in TWISTED's randomly collected dataset. We include only 3-crossing states with counts $\leq150$ and employ stratified sampling to ensure comprehensive coverage: from each frequency bin $[10k-9,10k]_{k=1}^{15}$, we randomly sample 2 states, yielding a total of 30 states. This approach allows us to systematically evaluate how both TWISTED-RL and TWISTED perform across varying levels of data availability. We hypothesize that TWISTED's performance will correlate positively with the number of times a target state was observed during training, while TWISTED-RL should demonstrate more robust generalization to rare states. Finally, we test on a 4-Crossing Set (4-Eval), which contains the top 10 most frequent states with 4 crossings in TWISTED's data. This set enables us to assess generalization to higher complexity knots beyond those encountered during training.

\begin{table}[t]
    \centering
    \renewcommand{\arraystretch}{1.2} 
    \resizebox{\columnwidth}{!}{%
    \begin{tabular}{l ccccc}
        \toprule
        & \multicolumn{4}{c}{\textbf{TWISTED-RL}} & \textbf{TWISTED} \\
        \cmidrule(lr){2-5} \cmidrule(lr){6-6}
        \textbf{Test Set} & \textbf{G} & \textbf{A} & \textbf{C} & \textbf{AC} & \textbf{Baseline} \\
        \midrule
        3-Easy        & 100 & 100 & 100          & 100 & 100 \\
        3-Medium      & 28  & 28  & \textbf{44}  & 26  & 0   \\
        3-Medium-1h   & 46  & 54  & \textbf{62}  & 34  & 0   \\
        3-Hard        & 38  & 20  & \textbf{46}  & 18  & 0   \\
        3-Eval        & 43  & 37  & \textbf{61}  & 35  & 0   \\
        4-Eval        & 24  & 24  & \textbf{26}  & 24  & 10  \\
        \bottomrule
    \end{tabular}%
    }
    \caption{Average Success rate (\%) of TWISTED vs TWISTED-RL variants across 5 different seeds.}
    \label{tab:comparison}
    \vspace{-2em}
\end{table}

\subsection{Results}
\label{sec:results}

We evaluate the four TWISTED-RL variants (G,A,C and AC) against TWISTED on five test sets of varying complexity. The metric used for comparison is the success rate over all seeds and states in each test set. Figure~\ref{fig:runtime}a presents a performance comparison between TWISTED and its TWISTED-RL variants on the 3-Easy test set, plotting the cumulative success rate (y-axis) against running time (x-axis). While both TWISTED and all TWISTED-RL variants achieve perfect performance with a 100\% success rate on the 3-Easy set, all TWISTED-RL variants demonstrate a substantial computational advantage. Notably, TWISTED-RL solves all problems approximately $3\times$ faster than the baseline TWISTED approach, highlighting the significant efficiency gains achieved through our method. Table~\ref{tab:comparison} presents the success rate on the 3 and 4 crossing number problem sets, where TWISTED-RL consistently outperforms TWISTED across all test sets. On 3-Medium, TWISTED-RL-C achieves the highest success rate (44\%), while TWISTED fails completely. Similarly, on 3-Hard, TWISTED-RL-C demonstrates the best performance (46\% success rate) compared to TWISTED's complete failure. When allowed extended runtime on 3-Medium (1 hour instead of 30 minutes), TWISTED-RL-C's success rate increases to 62\%, highlighting how our approach effectively scales with additional compute resources. In contrast, TWISTED shows no improvement with extended runtime. On the diverse 3-Eval test set, TWISTED-RL-C achieves the highest success rate (61\%), demonstrating its ability to tie knots across varying levels of complexity, while TWISTED again fails completely. On the 4-Eval set, representing complex knots beyond the training distribution, TWISTED-RL-C achieves a 26\% success rate, compared to TWISTED's 10\%, highlighting how our method generalizes to unseen knots. The results reveal a clear performance hierarchy among the TWISTED-RL variants, with TWISTED-RL-C consistently achieving the best performance across all test sets. TWISTED-RL-C outperforms other variants by substantial margins: achieving 44\% vs 28\% success rates on 3-Medium, 46\% vs 38\% on 3-Hard, and 61\% vs 43\% on 3-Eval compared to the second-best performing variant. This suggests that specialization by crossing number provides the most effective partitioning strategy for the knot-tying domain. TWISTED-RL-G and TWISTED-RL-A show comparable performance on most test sets, while TWISTED-RL-AC generally underperforms other variants. The anytime performance results in Fig.~\ref{fig:runtime}b primarily reflect how the evaluation sets were constructed, with each set intentionally containing knots of different (and increasingly challenging) difficulty tiers. Accordingly, success accumulates fastest on 3-Easy and slows progressively on 3-Medium/3-Hard/4-Eval, while 3-Eval sits in between due to its deliberately mixed difficulty distribution. These results establish TWISTED-RL as the new state-of-the-art in knot-tying without demonstrations, with TWISTED-RL-C emerging as the best overall variant. All variants outperform TWISTED by wide margins, with TWISTED-RL-C showing particularly strong advantages.

\section{Related Works}

Manipulating deformable objects presents varying levels of difficulty depending on the object class and task horizon \cite{matas2018sim}. In the context of rope manipulation, much prior work has focused either on learning from human demonstrations \cite{van2010superhuman, schulman2016learning, yan2020learning} or on solving short-horizon manipulation tasks, such as reshaping a rope via pick-and-place actions without forming knots \cite{wu2020learning, teng2022multidimensional}. In contrast, our work addresses long-horizon planning problems, specifically knot tying, in a fully demonstration-free setting. To cope with the complex and highly nonlinear dynamics of deformable objects, several approaches have relied on self-supervised learning to acquire predictive or inverse models of rope behavior \cite{sudry2023hierarchical, nair2017combining, yan2020self}. Differently, we use RL to avoid the data collection process inherited in self-supervised settings. RL has also been explored for rope manipulation \cite{lin2021softgym, han2017model, deng2022deep}, but these methods typically target different objectives than knot-tying.

\section{Limitations}
\label{sec:limitations}

Despite TWISTED-RL's advances over prior work, several important limitations remain. First, while simulation enables efficient policy learning, it introduces an inevitable sim2real gap. Our environment abstracts away several real-world challenges: we control a free-moving end-effector rather than a full robotic arm with kinematic constraints, and we assume perfect state estimation of the rope configuration. These simplifications are identical to those in the original TWISTED paper and would need to be addressed before deployment on physical systems. Additionally, despite substantial improvements, TWISTED-RL does not reach perfect success rates with medium and hard complexity states.  Addressing these challenges will require more sophisticated exploration strategies and potentially more expressive policy architectures to fully master the combinatorial complexity of the knot-tying domain. Finally, while TWISTED-RL's multi-step actions provide flexibility, they come with increased execution time compared to TWISTED's single-curve approach. 
For successful cases on the 3-Easy set, TWISTED-RL required an average of 2.5 curves per high-level action, compared to TWISTED's single curve. This creates a trade-off between success rate and execution time that practitioners should consider for time-sensitive applications.

\section{Conclusion}
\label{sec:conclusion}
This paper presents TWISTED-RL, a hierarchical framework that advances robotic knot-tying without human demonstrations. By decomposing the knot-tying problem into specialized agents dedicated to sub-problems, we overcome critical data-balancing challenges that hampered previous approaches. Our multi-step RL policies, conditioned on abstract topological actions rather than goal states, enable more flexible manipulation strategies and better generalization across diverse knots. Experimental results demonstrate TWISTED-RL's clear superiority over the previous state-of-the-art, extending capabilities to previously unsolvable complex knots with 3 and 4 crossings.


\section*{Appendix}

\subsection{Full Task Description}\label{appdx:extended-env} 
\subsubsection{Low-Level State Space}
\label{sec:low-level-state-space-appendix} 
We model the rope as a chain of $N$ links. Positions $P \subseteq \mathbb{R}^{3(N+1)}$: We store 3D coordinates $(x,y,z)$ of each of the $N+1$ joints in a global reference frame. Concretely, $p = (x_1, y_1, z_1, \dots, x_{N+1}, y_{N+1}, z_{N+1}) \in P$, where $(x_i,y_i,z_i)$ denotes the 3D position of the $i$-th joint. Orientations $O \subseteq \mathbb{R}^{2(N-1)+7}$: We decompose the rope’s orientation into a global transform with respect to the “middle” link (capturing its 3D position and a quaternion for its rotation) plus relative angles for the other links. Concretely, $o=(x_{\text{mid}}, y_{\text{mid}}, z_{\text{mid}}, q_w, q_x, q_y, q_z, a^p_1, a^y_1, \dots, a^p_{N-1}, a^y_{N-1}) \in O$ where the first 7 coordinates describe the position and orientation of the middle link. The remaining $2(N-1)$ coordinates correspond to yaw and pitch of each link relative to the previous link, omitting roll for simplicity.

\subsubsection{Low-Level Action Space}
\label{sec:low-level-action-space-appendix}
At the low level, actions occur in a continuous 3D workspace via a single robotic manipulator (a free-floating end-effector) as done in~\cite{sudry2023hierarchical}. 
We adopt \emph{curve-based-motion-primitives} (used by ~\cite{yan2020learning, sudry2023hierarchical}) to approximate the Reidmeister high-level actions. The parameters of a curve $c$ are: (i) Link to Grab $l$, an integer $l \in [1,N]$ specifying which rope link to grasp; (ii) Endpoint $(x,y)$, a planar target in the workspace to which the chosen link $l$ is moved, assuming the rope resides near the $z=0$ plane and within workspace limits; and (iii) Maximum Height $z_{max}$, the peak lift above the table during motion.

\subsubsection{Low-Level Simulation}
\label{sec:low-level-simulation-appendix}
We use the MuJoCo simulation from~\cite{sudry2023hierarchical} to simulate rope physics. A single robotic arm executes a “grab-lift-move-lower” trajectory, parameterized by a curve $c=(l, x, y, z_{\max}) \in C$. The low-level trajectory moves the hand from $(x_{start},y_{start},0)$ up to $(x_{start},y_{start},z_{max})$, then across to $(x,y,z_{max})$, and finally down to $(x,y,0)$. Once a low-level action action $c\in C$ is selected, the simulator applies it to the current configuration $q\in Q$, producing a new low-level state $q'$. The simulation terminates when the rope reaches zero velocity. Formally, we define this transition function as:
\(
f:Q\times C\longrightarrow Q
\).

\subsubsection{High-Level State Space}
\label{sec:high-level-state-space-appendix}
Let $Q$ be the low-level state space (Sec.~\ref{sec:low-level-state-space-appendix}). In this section, we will discuss the categorization of the low-level state space $Q$ to a high-level abstract state space $\mathcal{S}$. In rope manipulation, many different 3D arrangements may be equivalent from a topological perspective. For instance, two loops that look visually different could have the same topological states. Thus, an abstract representation better captures the \emph{essential} structure relevant to knot-tying while ignoring minor geometric differences.

\paragraph{P-Data as a Topological Representation}
\label{sec:p-data-appendix}
Following~\cite{yan2020learning, sudry2023hierarchical} we adopt a data structure called \emph{P-data}~\cite{pdata} to encode the topological state space $\mathcal{S}$. Each low-level configuration $q\in Q$ is mapped to its corresponding topological state $S \in \mathcal{S}$ via the mapping function:
\(
Top:Q \;\longrightarrow\; \mathcal{S}.
\)
P-data linearly organizes every intersection among rope segments, tracking for each intersection: (i) which pairs of segments are crossing (e.g., segment $k$ vs.\ segment $l$); (ii) which segment is on top, based on the larger $z$-value in that local region; and (iii) the sign of each crossing using a standard orientation convention. Every state holds its own unique P-data, and as the intricacy of the knot escalates, the representation of this P-data becomes progressively more comprehensive.

\paragraph{Crossing Number}
\label{sec:crossing-number-appendix}
While P-data fully encodes the class of the knot (i.e., which topological “type” the rope is in), the crossing number, defined by:
\(
\phi:\mathcal{S} \longrightarrow\mathbb{N}
\), 
provides a convenient measure of complexity. $\phi(S)$ is the minimal number of pairwise crossings in any valid arrangement of the rope belonging to topological class $S\in \mathcal{S}$. An unknotted loop has $\phi=0$, a simple trefoil knot has $\phi=3$, etc. The crossing number is a topological invariant: stretching or rotating the rope does not change the number of essential over/under crossings that define the knot’s “type.” Higher crossing numbers fully correlate with more intricate entanglements. Consequently, crossing number nicely approximates the difficulty of moving from one topological state to another.

\subsubsection{High-Level Action Space}
\label{sec:high-level-action-space-appendix}
In knot theory, Reidemeister moves~\cite{reidemeister1983knot} describe the fundamental ways a rope’s topological structure can be changed without cutting the rope. Formally, let $\mathcal{A}$ be the set of feasible Reidemeister moves. Reidemeister’s theorem states that any two topological states (or knots) can be transformed into one another by a finite sequence of Reidemeister moves $(A_0,A_1,\dots,A_m)$. To satisfy this, Reidemeister moves are reversible (Knot-Untying), but in our research, we will use only regular moves and not the corresponding reversible move, which are: Reidemeister I (R1) creates a single loop by pulling a free segment over or under itself; Reidemeister II (R2) adds a pair of intersections of opposite signs by pulling one segment on top of another; Reidemeister III (R3) rearranges two adjacent crossings without changing the overall crossing number; and Cross (C) introduces a new crossing by looping the rope’s head or tail over/under another segment. While all Reidemeister moves are theoretically important for transforming any knot into any other, we do not consider R3 in our current work, as they can be more challenging to achieve reliably in simulation (or require more specific rope-hand manipulations). Future studies may incorporate them to broaden the range of reachable knots. We thus define the set of high-level action types as $\{R1,R2,Cross\}$, and correspondingly the high-level action space $\mathcal{A}$, which is a combination of high-level action type and parameters. Our approach with R1, R2, and Cross can already generate a wide range of knots by incrementally increasing the crossing number, and empirically, many common knots are formed by repeatedly applying these three moves.

\subsubsection{High-Level Graph}
\label{sec:high-level-graph-appendix}

The high-level graph \(G=(\mathcal{S},\mathcal{A})\), where nodes \(\mathcal{S}\) correspond to topological states (Sec.~\ref{sec:high-level-state-space-appendix}), abstracting the underlying low-level state space \(s\) (Sec.~\ref{sec:low-level-state-space-appendix}), and edges \(\mathcal{A}\) denote Reidemeister moves (Sec.~\ref{sec:high-level-action-space-appendix}), abstracting the low-level action space \(Q\) (Sec.~\ref{sec:low-level-action-space-appendix}) is spanned by the high-level transition function $\mathcal{T}:\mathcal{S}\times \mathcal{A}\longrightarrow \mathcal{S}$. Notably, the number of topological states grows \emph{exponentially} with the crossing number \(\phi\). For instance, at \(\phi=0\), there is exactly one topological state -- the unknotted loop. As the complexity increases to \(\phi=1\), four distinct states emerge; at \(\phi=2\), there are dozens of states, and by \(\phi=3\), the graph already contains hundreds of possible states. This exponential growth illustrates the combinatorial complexity inherent to the knot-tying problem.

\subsection{Training Procedure}
\label{sec:training-appendix}

This section provides an overview of our training methodology. We detail the data flow of our method and comparing our data efficiency to TWISTED (Sec.~\ref{sec:data-flow}), and provide implementation details for the crossing number variants of our approach, explaining how policies are progressively trained for increasingly complex knots (Sec.~\ref{sec:additional-details-appendix}).

\subsubsection{Training Data Flow and Collection}
\label{sec:data-flow}
Our training scheme follows a purely self-supervised reinforcement learning approach with no external dependencies. The data flow operates as follows:

\paragraph{Data Collection Process} Training begins entirely from random policy rollouts for the first 2000 environment steps to ensure diverse initial exploration. After this bootstrap phase (as is common in RL), actions are generated by the learned agents themselves. Crucially, we do not use any data from TWISTED or other external sources. No hand-designed data collection strategies or expert demonstrations are incorporated into our training pipeline. This ensures that our approach relies solely on the agent's ability to learn from its own interactions with the environment.

\paragraph{Data Efficiency Comparison} To provide context on data requirements, we note that TWISTED used significantly more data than our approach. TWISTED required 1.67M data points from a dataset with $>$99\% drop rate, totaling $>$165M data points collected over 3 weeks. In contrast, our method uses only 2M environment steps per policy and completes training in at most 4 days, representing a substantial improvement in both data efficiency and computational requirements.

\subsubsection{Additional Details for Crossing Number Variants}
\label{sec:additional-details-appendix}

As described in Sec.~\ref{sec:high-level-planner}, variants that partition the high-level graph $G$ by crossing number create structured policy progressions. TWISTED-RL-AC, which trains specialized policies for each combination of crossing number and high-level action type, develops along the trajectory $\{\pi_0^{R1}, \pi_0^{R2}\} \longrightarrow \{\pi_1^{R1}, \pi_1^{R2}, \pi_1^{Cross}\} \longrightarrow \{\pi_2^{R1}, \pi_2^{R2}, \pi_2^{Cross}\} \dots$. In contrast, the TWISTED-RL-C variant, which trains a single specialized policy for each crossing number, follows the simpler progression $\pi_0 \longrightarrow \pi_1 \longrightarrow \pi_2 \dots$ The policies for crossing number 0 begin training with the unknotted rope configuration. As training progresses, the agent encounters additional low-level states with crossing number 0 that can serve as initial states for episodes (as described in Sec.~\ref{sec:selection}). However, for crossing numbers $> 0$, no suitable initial low-level states are available at the beginning of training, as we don't know which configurations belong to states with higher crossing numbers. To enable learning of policies with high crossing number without relying on complicated data collection schemes, we reuse previously encountered configurations; during training of policies for initial crossing number $l$, when we observe a transition from a configuration $q_l$ of crossing number $l$ to a configuration $q_k$ of crossing number $k$ where $k > l$, we save $q_k$ as a possible starting configuration for training policies with initial crossing number $k$. Later, when we start training policies for the initial crossing number $k$, we already have a diverse set of possible starting configurations available.


\bibliographystyle{IEEEtran}
\bibliography{IEEEabrv,root}

\begin{thebibliography}{10}
\providecommand{\url}[1]{#1}
\csname url@samestyle\endcsname
\providecommand{\newblock}{\relax}
\providecommand{\bibinfo}[2]{#2}
\providecommand{\BIBentrySTDinterwordspacing}{\spaceskip=0pt\relax}
\providecommand{\BIBentryALTinterwordstretchfactor}{4}
\providecommand{\BIBentryALTinterwordspacing}{\spaceskip=\fontdimen2\font plus
\BIBentryALTinterwordstretchfactor\fontdimen3\font minus \fontdimen4\font\relax}
\providecommand{\BIBforeignlanguage}[2]{{%
\expandafter\ifx\csname l@#1\endcsname\relax
\typeout{** WARNING: IEEEtran.bst: No hyphenation pattern has been}%
\typeout{** loaded for the language `#1'. Using the pattern for}%
\typeout{** the default language instead.}%
\else
\language=\csname l@#1\endcsname
\fi
#2}}
\providecommand{\BIBdecl}{\relax}
\BIBdecl

\bibitem{zhu2022challenges}
J.~Zhu, A.~Cherubini, C.~Dune, D.~Navarro-Alarcon, F.~Alambeigi, D.~Berenson, F.~Ficuciello, K.~Harada, J.~Kober, X.~Li \emph{et~al.}, ``Challenges and outlook in robotic manipulation of deformable objects,'' \emph{IEEE Robotics \& Automation Magazine}, vol.~29, no.~3, pp. 67--77, 2022.

\bibitem{gu2023survey}
F.~Gu, Y.~Zhou, Z.~Wang, S.~Jiang, and B.~He, ``A survey on robotic manipulation of deformable objects: Recent advances, open challenges and new frontiers,'' \emph{arXiv preprint arXiv:2312.10419}, 2023.

\bibitem{wang2010suturing}
H.~Wang, S.~Wang, J.~Ding, and H.~Luo, ``Suturing and tying knots assisted by a surgical robot system in laryngeal mis,'' \emph{Robotica}, vol.~28, no.~2, pp. 241--252, 2010.

\bibitem{6631059}
K.~Nishibori and K.~Nishibori, ``Automation of tying task on tie-dyeing of traditional craft by robots,'' in \emph{2013 IEEE International Conference on Robotics and Automation}, 2013, pp. 3451--3456.

\bibitem{shivakumar2023sgtm}
K.~Shivakumar, V.~Viswanath, A.~Gu, Y.~Avigal, J.~Kerr, J.~Ichnowski, R.~Cheng, T.~Kollar, and K.~Goldberg, ``Sgtm 2.0: Autonomously untangling long cables using interactive perception,'' in \emph{2023 IEEE International Conference on Robotics and Automation (ICRA)}.\hskip 1em plus 0.5em minus 0.4em\relax IEEE, 2023, pp. 5837--5843.

\bibitem{sudry2023hierarchical}
M.~Sudry, T.~Jurgenson, A.~Tamar, and E.~Karpas, ``Hierarchical planning for rope manipulation using knot theory and a learned inverse model,'' in \emph{Conference on Robot Learning}.\hskip 1em plus 0.5em minus 0.4em\relax PMLR, 2023, pp. 1596--1609.

\bibitem{reidemeister1983knot}
K.~Reidemeister, \emph{Knot theory}.\hskip 1em plus 0.5em minus 0.4em\relax BCS Associates, 1983.

\bibitem{712192}
R.~Sutton and A.~Barto, ``Reinforcement learning: An introduction,'' \emph{IEEE Transactions on Neural Networks}, vol.~9, no.~5, pp. 1054--1054, 1998.

\bibitem{sac}
T.~Haarnoja, A.~Zhou, P.~Abbeel, and S.~Levine, ``Soft actor-critic: Off-policy maximum entropy deep reinforcement learning with a stochastic actor,'' in \emph{International conference on machine learning}.\hskip 1em plus 0.5em minus 0.4em\relax Pmlr, 2018, pp. 1861--1870.

\bibitem{pdata}
J.~Takamatsu, T.~Morita, K.~Ogawara, H.~Kimura, and K.~Ikeuchi, ``Representation for knot-tying tasks,'' \emph{IEEE Transactions on Robotics}, vol.~22, no.~1, pp. 65--78, 2006.

\bibitem{goexplore}
\BIBentryALTinterwordspacing
A.~Ecoffet, J.~Huizinga, J.~Lehman, K.~O. Stanley, and J.~Clune, ``Go-explore: a new approach for hard-exploration problems,'' \emph{CoRR}, vol. abs/1901.10995, 2019. [Online]. Available: \url{http://arxiv.org/abs/1901.10995}
\BIBentrySTDinterwordspacing

\bibitem{mujoco}
E.~Todorov, T.~Erez, and Y.~Tassa, ``Mujoco: A physics engine for model-based control,'' in \emph{2012 IEEE/RSJ International Conference on Intelligent Robots and Systems}, 2012, pp. 5026--5033.

\bibitem{matas2018sim}
J.~Matas, S.~James, and A.~J. Davison, ``Sim-to-real reinforcement learning for deformable object manipulation,'' in \emph{Conference on Robot Learning}.\hskip 1em plus 0.5em minus 0.4em\relax PMLR, 2018, pp. 734--743.

\bibitem{van2010superhuman}
J.~Van Den~Berg, S.~Miller, D.~Duckworth, H.~Hu, A.~Wan, X.-Y. Fu, K.~Goldberg, and P.~Abbeel, ``Superhuman performance of surgical tasks by robots using iterative learning from human-guided demonstrations,'' in \emph{2010 IEEE International Conference on Robotics and Automation}.\hskip 1em plus 0.5em minus 0.4em\relax IEEE, 2010, pp. 2074--2081.

\bibitem{schulman2016learning}
J.~Schulman, J.~Ho, C.~Lee, and P.~Abbeel, ``Learning from demonstrations through the use of non-rigid registration,'' in \emph{Robotics Research}.\hskip 1em plus 0.5em minus 0.4em\relax Springer, 2016, pp. 339--354.

\bibitem{yan2020learning}
M.~Yan, G.~Li, Y.~Zhu, and J.~Bohg, ``Learning topological motion primitives for knot planning,'' in \emph{2020 IEEE/RSJ International Conference on Intelligent Robots and Systems (IROS)}.\hskip 1em plus 0.5em minus 0.4em\relax IEEE, 2020, pp. 9457--9464.

\bibitem{wu2020learning}
Y.~Wu, W.~Yan, T.~Kurutach, L.~Pinto, and P.~Abbeel, ``Learning to manipulate deformable objects without demonstrations,'' in \emph{16th Robotics: Science and Systems, RSS 2020}.\hskip 1em plus 0.5em minus 0.4em\relax MIT Press Journals, 2020.

\bibitem{teng2022multidimensional}
Y.~Teng, H.~Lu, Y.~Li, T.~Kamiya, Y.~Nakatoh, S.~Serikawa, and P.~Gao, ``Multidimensional deformable object manipulation based on dn-transporter networks,'' \emph{IEEE Transactions on Intelligent Transportation Systems}, 2022.

\bibitem{nair2017combining}
A.~Nair, D.~Chen, P.~Agrawal, P.~Isola, P.~Abbeel, J.~Malik, and S.~Levine, ``Combining self-supervised learning and imitation for vision-based rope manipulation,'' in \emph{2017 IEEE international conference on robotics and automation (ICRA)}.\hskip 1em plus 0.5em minus 0.4em\relax IEEE, 2017, pp. 2146--2153.

\bibitem{yan2020self}
M.~Yan, Y.~Zhu, N.~Jin, and J.~Bohg, ``Self-supervised learning of state estimation for manipulating deformable linear objects,'' \emph{IEEE robotics and automation letters}, vol.~5, no.~2, pp. 2372--2379, 2020.

\bibitem{lin2021softgym}
X.~Lin, Y.~Wang, J.~Olkin, and D.~Held, ``Softgym: Benchmarking deep reinforcement learning for deformable object manipulation,'' in \emph{Conference on Robot Learning}.\hskip 1em plus 0.5em minus 0.4em\relax PMLR, 2021, pp. 432--448.

\bibitem{han2017model}
H.~Han, G.~Paul, and T.~Matsubara, ``Model-based reinforcement learning approach for deformable linear object manipulation,'' in \emph{2017 13th IEEE Conference on Automation Science and Engineering (CASE)}.\hskip 1em plus 0.5em minus 0.4em\relax IEEE, 2017, pp. 750--755.

\bibitem{deng2022deep}
Y.~Deng, C.~Xia, X.~Wang, and L.~Chen, ``Deep reinforcement learning based on local gnn for goal-conditioned deformable object rearranging,'' in \emph{2022 IEEE/RSJ International Conference on Intelligent Robots and Systems (IROS)}.\hskip 1em plus 0.5em minus 0.4em\relax IEEE, 2022, pp. 1131--1138.

\end{thebibliography}

\end{document}